\documentclass{article}
\usepackage{spconf,amsmath,graphicx}
\usepackage{algorithm}
\usepackage[noend]{algpseudocode}

\title{Real-time tracker with fast recovery from target loss}
\name{Alessandro Bay$^\star$, Panagiotis Sidiropoulos$^{\star\dagger}$, Eduard Vazquez$^\star$, Michele Sasdelli$^{\star\sharp}$}
\address{$^\star$Cortexica Vision System Ltd, London, UK\\
$^\dagger$Space and Climate Physics Department, University College London, UK\\
$^\sharp$Australian Institute for Machine Learning, The University of Adelaide, Australia}
\begin{document}
%
\maketitle
\begin{abstract}
In this paper, we introduce a variation of a state-of-the-art real-time tracker (CFNet), which adds to the original algorithm robustness to target loss without a significant computational overhead. The new method is based on the assumption that the feature map can be used to estimate the tracking confidence more accurately. When the confidence is low, we avoid updating the object's position through the feature map; instead, the tracker passes to a single-frame failure  mode, during which the patch's low-level visual content is used to swiftly update the object's position, before recovering from the target loss in the next frame. The experimental evidence provided by evaluating the method on several tracking datasets validates both the theoretical assumption that the feature map is associated to tracking confidence, and that the proposed implementation can achieve target recovery in multiple scenarios, without compromising the real-time performance.
\end{abstract}


\section{Introduction}\label{sec:intro}

The recent advances in deep learning research has significantly improved the state-of-the-art in a multitude of computer vision and multimedia processing applications such as object detection \cite{girshick2014rich}, 
semantic segmentation \cite{long2015fully}, action recognition \cite{simonyan2014two}, etc. However, in tracking applications deep learning initially struggled to exhibit performance improvements \cite{HNam16}. Only recently deep learning has significantly improved the state-of-the-art, both for real-time \cite{bertinetto2016fully,valmadre2017end} and non-real-time tracking \cite{SYun17,MDanelljan16}. 

A methodological modification that greatly benefitted deep-learning trackers was the replacement of the off-line training on large classification datasets with either on-line
training (e.g. ECO \cite{MDanelljan17}) or with off-line training on an image retrieval (and not classification) setup (e.g. CFNet \cite{valmadre2017end}). Each of these two classes of algorithms achieve state-of-the-art performance in one of the two main video tracking sub-categories. More specifically, algorithms using on-line training are optimal in non-real-time tracking and algorithms using image retrieval training in real-time tracking. 

However, several significant challenges remain unresolved, including the recovery from target loss \cite{RTao17} and the sensitivity to distractors \cite{LWang15}. This work aims to reduce the sensitivity of the state-of-the-art CFNet tracker from target loss without significantly reducing its speed. The difficulty of tracker recovery originates from its main design principle, i.e. its ability to accumulate correct object positions for a substantial amount of time. This ability becomes a severe flaw in the cases that the tracker would temporarily lose the position of the object due to an abrupt camera movement, an unexpected and abrupt object movement, a technical problem, a compression error, etc. Once the sampling drift \cite{RTao17} causes the tracker bounding box to not intersect with the object, the tracker capability to accumulate correct retrievals ''locks'' the object in the background with little possibility of recovery.

The tracking-through-similarity paradigm that is adopted by CFNet \cite{valmadre2017end} provides a mechanism that can be used for on-line identification of target loss. More specifically, in this work we follow the common (e.g. \cite{DLowe04}) nearest neighbour distance ratio (NNDR) to declare ambiguous or unambiguous tracking updates. If the CFNet output is declared ambiguous the tracker is entering a 1-frame ``failure mode'', during which (1) the tracker is updated using a backup tracker and (2) the search area is doubled to facilitate the object redetection in the next frame. The implementation of the architecture introduced in this paper is named CFNet-FTLR (Fast Target Loss Recovery) and uses a simple low-level visual feature correlation scheme as a backup tracker. Apart from the original architecture, the main contributions of the paper include (1) the ambiguity measure that estimates the confidence on the tracker output, (2) the enlargement of the search area when the tracker is on failure mode, (3) the use of a simple low-level representation as a short-term backup tracker, and (4) an improved running average equation for the query model.

\section{Related work}\label{sec:relworks}

Due to its major significance, the tracking of moving objects in videos has a long history of research and development. The matureness of the domain partially explains the failure of deep-learning trackers to outperform ``classical'' trackers \cite{HNam16}. Perhaps a more important reason is the inherent characteristics of the tracking setup that undermined a straightforward transfer of deep-learning techniques to this task \cite{DZhang17}. More specifically, (1) maximising heatmaps corresponding to semantic classes is not necessarily the optimal strategy to locate a specific object \cite{CMa15} (especially in the presence of distractors), (2) off-line training is hampered by the lack of large-volume annotated datasets for video tracking, and (3) on-line training is prohibitively slow for applications requiring real-time tracking, especially if the model update is conducted in each and every frame \cite{MDanelljan17,RTao17}.

On the other hand, the small temporal window between two consecutive frames implies a visual similarity of the tracked object. Based on this rationale, a tracking-by-similarity tracker has been introduced  \cite{bertinetto2016fully}, in which similarity is learned through a Siamese deep network that is trained offline, while the localisation is conducted through a correlation filter \cite{valmadre2017end}. This method achieved state-of-the-art real-time performance, despite its simple architecture.

One of the main issues with such a tracker is the sampling drift \cite{RTao17}, from which the tracker often fails to recover, that occurs in the case of abrupt object motion, camera motion, etc. Two main solutions have been recently proposed: (1) the heatmap generated in the final stage of the algorithm can incorporate the localisation of the ambiguity by not exhibiting a clear peak and in this case it rejects the position update \cite{JValmadre18}, and (2) the model drift can be avoided through a global similarity search (in the whole frame) every $N$ frames ($N$ being a hard-coded constant parameter) \cite{RTao17}.

In this work, we carefully combine the above solutions in order to optimise both the accuracy and the computational time. The localisation ambiguity is detected using the Nearest Neighbour Distance Ratio (NNDR), instead of through a peak comparison to a hard-coded threshold. After rejecting an ambiguous position update, tracking is conducted in this frame using correlation of low-level patch representations as a backup tracker. This is inspired by the recent improvements on tracker performance that was achieved by including the first network layers, which model the low-level image content \cite{MDanelljan16,MDanelljan17}. Finally, in the next frame, the search window is expanded, but without covering the whole frame, as in \cite{RTao17}, so as to reduce the computational cost. 

\section{CFNet-FTLR  tracker}\label{sec:HnS}
\begin{figure}[htb]
\centering{
\includegraphics[scale=0.45]{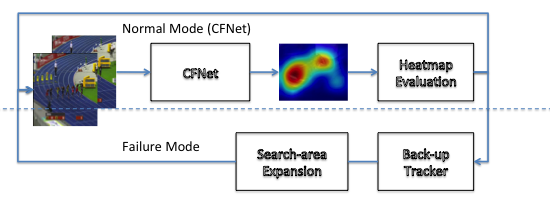}
\caption{The architecture of the CFNet-FTLR tracker}
\label{fig:architecture}}
\end{figure}

The main concept of our CFNet-Fast Target Loss Recovery (CFNet-FTLR, Figure \ref{fig:architecture}) tracker lies in the assumption that the CFNet feature map can be used to evaluate the confidence on the bounding box update. The confidence is modelled through the ratio of the two most dominant peaks, which are estimated by projecting the 3D map onto $yz$ and $xz$ planes, before being differentiated twice in order to identify the local maxima, which determine the two most dominant peaks. Since CFNet is estimating patch similarity, these peaks correspond to the nearest neighbour and the second nearest neighbour of the current bounding box. 

Their ratio (i.e. NNDR) is thresholded to evaluate the confidence on the tracker output. More specifically, if the ratio is above the confidence threshold, then the tracker output is considered safe. Therefore, the top peak is followed to update the object position and the tracking continues following the standard CFNet algorithm in the next frame. On the other hand, if the ratio is below the confidence threshold, the tracker output is considered ambiguous and the tracker passes to failure mode.

The following measures are taken during the time that the tracker is on failure mode: (1) CFNet output is not used to update the object position (the top peak position is ignored), (2) the object position is updated following the estimation of the backup tracker, and (3) in the following frame  the object is searched in an area which is wider than the original one.

The backup tracker is based on correlating Census-transformed \cite{RZabih94} image patches.
Census transform is a simple and powerful low-level representation of the image content that holds a set of desirable features: (1) it has linear computational complexity, (2) it preserves the object edges, (3) it is robust to radiometric differences that are not uncommon in videos \cite{HHirschmuller09}, and (4) it generates robust optical flow estimations \cite{DHafner13}. Moreover, from a deep-learning point of view, Census transform could be considered as a hand-crafted filter of the first layer of a neural network, i.e. it fits to the recent results implying that low-level visual information can contribute to the overall tracker performance. The 8-bit binary strings that the Census transform generates for each pixel are converted into $4$ decimal numbers by iteratively applying a circular shift of $2$ positions before conversion. The result is correlated and the position of the maximum value is followed to update the object bounding box.

Additionally, the standard approach used in CFNet \cite{bertinetto2016fully} to create the query model from the previously seen feature map is a simple running average:
\begin{equation}\label{Eq:1}
Q_1 = F_1, \qquad
Q_n = (1-\alpha)Q_{n-1} + \alpha F_n,
\end{equation}
where $Q_n$ is the $n$-th query model, $F_n$ is the $n$-th feature map, and the update factor $\alpha$ is empirically set to $\alpha=0.005$. 

An unintentional result of Eq.\ \eqref{Eq:1} is that during the initial video frames (in which case $n \not\gg 1/\alpha$), $Q_n$ is dominated by the first feature map $F_1$. For example, if $\alpha=0.005$ and the frame ratio is $25$ fps, then $8$ seconds after the video started $F_1$ would determine $37\%$ of the generated query feature map. In the general case, such a strong dependence from $F_1$ is expected to be suboptimal, especially in practical applications, where the tracked object would be initialised from an arbitrary pose, and frequently from a frame of poor visible quality.

Therefore, it is suggested to reduce the dependence from the first frame, by replacing Eq.\ \eqref{Eq:1} with the following running average formula:
\begin{equation}\label{Eq:2}
Q_n = (1 - 0.5/n - \alpha)Q_{n-1} + (0.5/n + \alpha)F_n.
\end{equation}
Eqs.\ \eqref{Eq:1} and \eqref{Eq:2} converge asymptotically. Their main difference is that Eq.\ \eqref{Eq:2} generates a ``smooth average'' (SA) over the first frames, by creating a bootstrap model which uses a large number of initial frames ($\sim \frac{1}{5\alpha}$), instead of a single frame. This improves significantly the performance in the more challenging TRE evaluation, a benchmark that measures the robustness of the tracker on the initial object pose. The introduced algorithm (CFNet-FTLR{\_}SA) is analytically presented in Algorithm \ref{alg:hideandseek}. 

\begin{algorithm} \footnotesize
\caption{CFNet-FTLR{\_}SA tracker($video$, $b_0$, $NNDR$, $DefaultImageArea$)}
\label{alg:hideandseek}
\begin{algorithmic}[1]
\State (Initialisation) $ImageArea=DefaultImageArea$ 
\State Select the next frame and crop a patch of size equal to $ImageArea$ around object region 
\State Resize the patch to fit the input size of the Neural Network (NN) 
\State Forward pass the patch to the NN and estimate the feature map 
\State Estimate the correlation map from the feature map and the existing running average
\State Estimate the two highest peaks in the correlation map, $P_1$ and $P_2$
 \If {$P_1/P_2>NNDR$} \\
  \quad(a) update the position \\
  \quad(b) the running average using Eq.\ \eqref{Eq:2} \\
  \quad(c) put $ImageArea=DefaultImageArea$
 \Else \\
 \quad (a) update the position after running the backup tracker \\
 \quad (b) put $ImageArea=2*DefaultImageArea$  
\EndIf
\If {this is the last video frame} \Return
\EndIf
\end{algorithmic}
\end{algorithm}

\begin{table*}[htbp]
\centering{
\begin{tabular}{l | c | c c c c | c c c c | c c c c}
& & \multicolumn{4}{c}{OTB-2013} & \multicolumn{4}{|c|}{OTB-50} & \multicolumn{4}{c}{OTB-100} \\
& & \multicolumn{2}{c}{OPE} & \multicolumn{2}{c}{TRE} & \multicolumn{2}{|c}{OPE} & \multicolumn{2}{c|}{TRE} & \multicolumn{2}{c}{OPE} & \multicolumn{2}{c}{TRE} \\
Method & fps & IoU & prec & IoU & prec & IoU & prec & IoU & prec & IoU & prec & IoU & prec \\
\hline
CFNet \cite{valmadre2017end} & 71.1 & 57.0 & 74.1 & 59.9 & 76.1 & 49.4 & 64.3 & 52.9 & 69.0 & 55.7 & 71.6 & 58.2 & 73.7 \\
CFNet-FTLR{\_}0 & 64.2 & 57.4 & 74.4 & 60.2 & 76.5 & 49.1 & 64.1 & 53.0 & 69.2 & 54.6 & 70.3 & 58.0 & 73.4 \\
CFNet-FTLR{\_}1 & 64.0 & 57.9 & 75.2 & 60.2 & 76.5 & 48.9 & 63.8 & 53.0 & 69.0 & 54.8 & 69.9 & 58.1 & 73.6 \\
CFNet-FTLR  & 61.6 & \bf{60.0} & {\bf78.2} & 60.6 & 76.9 & 51.3 & 68.1 & 53.9 & 70.7 & {\bf57.3} & {74.1} & 58.7 & 74.4 \\
CFNet-FTLR{\_}SA & 62.3 & 58.7 & 76.8 & {\bf61.9} & \bf{79.8} & \bf{51.5} & {\bf68.5} & \bf{55.0} & {\bf73.2} & 57.1 & \textbf{74.6} & {\bf60.3} & {\bf77.8} \\
\hline
CFNet-FTLR{\_}GT & 65.0 & 62.7 & 83.6 & 65.3 & 84.6 & 54.6 & 74.6 & 58.7 & 79.3 & 59.3 & 78.2 & 62.5 & 80.8 \\
\end{tabular}
\caption{The results on OTB dataset for our baseline (CFNet) along with the $4$ tested variations of FTLR: FTLR0, FTLR1, FTLR and FTLR with smooth average (FTLR{\_}SA). The best performance is highlighted in bold.}
\label{tab:resOTB}}
\end{table*}

\section{Experimental Results}\label{sec:res}
\subsection{Implementation Details and details}\label{subsec:implement}
Using the CFNet implementation provided by the authors, we have gradually augmented it with more modifications in order to confirm the contribution of all parts of the algorithm. Therefore, apart from the final algorithm CFNet-FTLR{\_}SA, as well as the original CFNet method, there are three more variations experimentally evaluated: (1) CFNet-FTLR\_0, in which the position of the bounding box is not updated until the NNDR is above the confidence threshold, but the search area for the subsequent frames is twice the original one, (2) CFNet-FTLR\_1, in which the new position is extrapolated by the position of the object in the past two frames using bilinear interpolation, and (3) CFNet-FTLR, in which the smooth average is not included in the algorithm.

We additionally tested the validity of our basic assumption, i.e. that the feature map NNDR is associated with tracking ambiguity, by evaluating also a ``theoretic tracker'', which achieves the best performance of a tracker using the normal mode/failure mode architecture. This ``theoretic tracker'' is modelled by replacing the backup tracker with a ``perfect'' one that always returns the ground truth, i.e. by feeding the ground truth as the new object position whenever an ambiguity is identified. The performance obtained by this ``tracker'' (named CFNet-FTLR{\_}GT) is the upper limit that can be reached by the introduced architecture, while CFNet-FTLR{\_}SA is a first implementation towards this direction.

For evaluating our algorithm, we use three tracking benchmarks: (a) OTB-100, OTB-50 and OTB-2013 \cite{valmadre2017end} datasets created from the Object Tracking Benchmark (OTB) \cite{WuLimYang13}, (b) UAV-123 \cite{mueller2016benchmark}, which contains videos collected from from low-altitude Unmanned Aerial Vehicles, and (c) a unified tracking benchmark on drone platforms (DTB-70) \cite{drone-tracking}. For all benchmarks we report the results for One-Pass Evaluation (OPE) and Temporal Robustness Evaluation (TRE), following the standard literature approach (e.g. \cite{LWang15}), while the tracker speed (in fps) is also reported for each method.

All of the calculations were performed with MATLAB R2017a, MatConvNet 1.0-beta25, Cuda-8.0, Cudnn-5.1, on a i7-6800K CPU @ 3.40GHz $\times$ 12 workstation with 32 GB RAM and a single nVidia GeForce GTX 1080Ti GPU.

\subsection{Results}
Tables \ref{tab:resOTB}, \ref{tab:resUAV} and \ref{tab:resDTB} summarise the performance for both OPE and TRE evaluation, comparing our methods with the subset of literature algorithms that also perform real-time on-line tracking.
Note that the Tables include the theoretic boundary of the performance, which is modelled by CFNet-FTLR{\_}GT, while for the OTB dataset, CFNet-FTLR{\_}0 and CFNet-FTLR{\_}1 are also evaluated.

%
%

A first conclusion that can be reached from these results is related to the validity of the assumption that the object ambiguity is correlated with a feature map presenting multiple peaks. More specifically, CFNet-FTLR{\_}GT implies that
a CFNet variation exploiting this property using a fast and accurate backup tracker could achieve accuracy and precision improvement as high as $13\%$ and $24.5\%$, respectively (OPE, DTB-70 dataset). The median accuracy and precision absolute improvements are $5.7\%$ and $10.3\%$, respectively, a substantial increase which would bring CFNet near the performance currently achieved only by non-real-time trackers.

On the other hand, comparing CFNet-FTLR\_SA with the original CFNet algorithm shows that our variation consistently outperforms the original CFNet at a computational overhead (CFNet-FTLR\_SA runs at $62.3$ fps while CFNet at $71.1$ fps) that for most applications would be considered negligible. The median accuracy and precision increase is $2.1\%$ and $4.1\%$, respectively, while for the OPE evaluation in the DTB-70 dataset it is $3.6\%$ and $6.6\%$. 

\begin{table}[t]
\centering{
\begin{tabular}{l | c c c c }
& \multicolumn{2}{c}{OPE} & \multicolumn{2}{c}{TRE} \\
Method & IoU & prec & IoU & prec \\
\hline
KCF \cite{henriques2015high} & 33.1 & 52.3 & -- & -- \\
DSST \cite{danelljan2014accurate} & 35.6 & 58.6 & -- & -- \\
CFNet \cite{valmadre2017end} & 47.0 & 66.5 & 52.4 & 72.2 \\
CFNet-FTLR  & 47.2 & 67.0 & \textbf{53.3} & 73.7 \\
CFNet-FTLR{\_}SA & \textbf{47.6} & \textbf{67.6} & 52.9 & \textbf{73.8} \\
\hline
CFNet-FTLR{\_}GT & 54.9 & 80.7 & 59.2 & 84.5 \\
\end{tabular}
\caption{The results on UAV-123 dataset for our baseline (CFNet) along with two FTLR variations. Two more literature, on-line and real-time, methods are included in the comparison. The best performance is highlighted in bold.}
\label{tab:resUAV}}
\end{table}

By comparing the achieved improvement with the theoretic upper boundary, it can be deduced that CFNet-FTLR\_SA exploits approximately $40\%$ of the additional accuracy and precision that the use of NNDR allows. This is far from optimal, but it should be contrasted with simple solutions such as CFNet-FTLR\_0 and CFNet-FTLR\_1, which completely fail to improve the results. As a matter of fact, both CFNet-FTLR\_0 and CFNet-FTLR\_1 median accuracy and precision is almost identical to CFNet, exhibiting performance deterioration for half of the evaluations that have been tested.

Tables \ref{tab:resOTB}, \ref{tab:resUAV} and \ref{tab:resDTB} include not only CFNet but also all real-time on-line trackers that we could find to report results on the same datasets. In all datasets included in this work, the CFNet-FTLR\_SA variation exhibits state-of-the-art performance, since it outperforms all included on-line real-time literature techniques. On the other hand, if this method is compared with non-real-time state-of-the-art trackers, 
it seems that it exhibits worse performance than the best current non-real-time trackers (such as MDNet \cite{HNam16} in the DTB-70 dataset). However, the performance gap between MDNet and the original CFNet was $6.2\%$ and $11.1\%$ in success and precision, respectively. By using CFNet-FTLR{\_}SA, instead, this performance gap (which can be viewed as the performance gap between non-real-time and real-time trackers) is reduced to $2.6\%$ and $4.5\%$.

\begin{table}[t]
\centering{
\begin{tabular}{l | c c c c }
& \multicolumn{2}{c}{OPE} & \multicolumn{2}{c}{TRE} \\
Method & IoU & prec & IoU & prec \\
\hline
DSST \cite{danelljan2014accurate} & 26.4 & 40.2 & -- & -- \\
KCF \cite{henriques2015high} & 28.0 & 46.8 & -- & -- \\
CFNet \cite{valmadre2017end} & 39.4 & 57.9 & 48.1 & 67.2 \\
CFNet-FTLR  & 41.2 & 61.3 & 49.1 & 68.7 \\
CFNet-FTLR{\_}SA & \textbf{43.0} & \textbf{64.5} & \textbf{50.7} & \textbf{72.2} \\
\hline
CFNet-FTLR{\_}GT & 52.4 & 82.2 & 56.3 & 81.9 \\
\end{tabular}
\caption{The results on DTB-70 dataset for our baseline (CFNet) along with two FTLR variations. Two more literature, on-line and real-time, methods are included in the comparison. The best performance is highlighted in bold.}
\label{tab:resDTB}}
\end{table}

Finally, the comparison between CFNet-FTLR\_SA and CFNet-FTLR validates the analysis conducted in the previous section. CFNet-FTLR\_SA outperforms CFNet-FTLR in all but four (OPE and TRE) evaluations, while it shows a slightly better performance in TRE evaluation than in OPE evaluation. This is aligned with the dependence of the TRE from the object viewing angle (in the original frame), thus signifying that in most practical applications (in which the object viewing angle in the first frame is not generally known) CFNet-FTLR\_SA should be preferred.

\section{Conclusions}\label{sec:concl}
In this work, we examined the hypothesis that in tracking-through-similarity algorithms the output heatmap that determines the object position in the next frame could be used to detect possible sampling drift. Moreover, we introduced an algorithm based on this hypothesis that employs a 1-frame backup tracker to temporarily update the object position, thus allowing the tracker to recover from target loss in subsequent frames. The experimental results in three distinct datasets confirm the potential of the introduced method.

{
\bibliographystyle{IEEEbib}

\begin{thebibliography}{10}

\bibitem{girshick2014rich}
Ross Girshick, Jeff Donahue, Trevor Darrell, and Jitendra Malik,
\newblock ``Rich feature hierarchies for accurate object detection and semantic
  segmentation,''
\newblock in {\em Proceedings of the IEEE conference on computer vision and
  pattern recognition}, 2014, pp. 580--587.

\bibitem{long2015fully}
Jonathan Long, Evan Shelhamer, and Trevor Darrell,
\newblock ``Fully convolutional networks for semantic segmentation,''
\newblock in {\em Proceedings of the IEEE conference on computer vision and
  pattern recognition}, 2015, pp. 3431--3440.

\bibitem{simonyan2014two}
Karen Simonyan and Andrew Zisserman,
\newblock ``Two-stream convolutional networks for action recognition in
  videos,''
\newblock in {\em Advances in neural information processing systems}, 2014, pp.
  568--576.

\bibitem{HNam16}
H.~Nam and B.~Han,
\newblock ``Learning multi-domain convolutional neural networks for visual
  tracking,''
\newblock in {\em Proceedings of the IEEE conference on computer vision and
  pattern recognition}, 2016.

\bibitem{bertinetto2016fully}
Luca Bertinetto, Jack Valmadre, Joao~F Henriques, Andrea Vedaldi, and Philip~HS
  Torr,
\newblock ``Fully-convolutional siamese networks for object tracking,''
\newblock in {\em European conference on computer vision}. Springer, 2016, pp.
  850--865.

\bibitem{valmadre2017end}
Jack Valmadre, Luca Bertinetto, Jo{\~a}o Henriques, Andrea Vedaldi, and
  Philip~HS Torr,
\newblock ``End-to-end representation learning for correlation filter based
  tracking,''
\newblock in {\em Computer Vision and Pattern Recognition (CVPR), 2017 IEEE
  Conference on}. IEEE, 2017, pp. 5000--5008.

\bibitem{SYun17}
S.~Yun, J.~Choi, Y.~Yoo, K.~Yun, and J.~Y. Choi,
\newblock ``Action-decision networks for visual tracking with deep
  reinforcement learning,''
\newblock in {\em Proceedings of the IEEE conference on computer vision and
  pattern recognition}, 2017.

\bibitem{MDanelljan16}
M.~Danelljan, A.~Robinson, F.~Khan, and M.~Felsberg,
\newblock ``Beyond correlation filters: Learning continuous convolution
  operators for visual tracking,''
\newblock in {\em IEEE European conference on computer vision}, 2016.

\bibitem{MDanelljan17}
M.~Danelljan, G.~Bhat, F.~S. Khan, and M.~Felsberg,
\newblock ``Eco: Efficient convolution operators for tracking,''
\newblock in {\em Proceedings of the IEEE conference on computer vision and
  pattern recognition}, 2017, pp. 6931--6939.

\bibitem{RTao17}
R.~Tao, E.~Gavves, and A.~W.M. Smeulders,
\newblock ``Tracking for half an hour,''
\newblock in {\em arXiv preprint arXiv:1711.10217}, 2017.

\bibitem{LWang15}
L.~Wang, W.~Ouyang, X.~Wang, and H.~Lu,
\newblock ``Visual tracking with fully convolutional networks,''
\newblock in {\em IEEE International conference on computer vision}, 2015.

\bibitem{DLowe04}
D.~Lowe,
\newblock ``Distinctive image features from scale-invariant keypoints,''
\newblock {\em Int. Journal of Computer Vision}, vol. 60, no. 2, pp. 91--110,
  2004.

\bibitem{DZhang17}
D.~Zhang, H.~Maei, X.~Wang, and Y.-F. Wang,
\newblock ``Deep reinforcement learning for visual object tracking in videos,''
\newblock in {\em arXiv preprint arXiv:1701.08936}, 2017.

\bibitem{CMa15}
C.~Ma, J.-B. Huang, X.~Yang, and M.-H. Yang,
\newblock ``Hierarchical convolutional features for visual tracking,''
\newblock in {\em IEEE International conference on computer vision}, 2015.

\bibitem{JValmadre18}
J.~Valmadre, L.~Bertinetto, J.~F. Henriques, R.~Tao, A.~Vedaldi, A.~Smeulders,
  P.~Torr, and E.~Gavves,
\newblock ``Long-term tracking in the wild: A benchmark,''
\newblock in {\em arXiv preprint arXiv:1803.09502}, 2018.

\bibitem{RZabih94}
R.~Zabih and J.~Woodfill,
\newblock ``Non-parametric local transforms for computing visual
  correspondence,''
\newblock in {\em IEEE European conference on computer vision}, 1994, pp.
  151--158.

\bibitem{HHirschmuller09}
H.~Hirschmuller and D.~Scharstein,
\newblock ``Evaluation of stereo matching costs on images with radiometric
  differences,''
\newblock {\em IEEE Transactions on Pattern Analysis and Machine Intelligence},
  vol. 31, no. 9, pp. 1582--1599, 2009.

\bibitem{DHafner13}
D.~Hafner, O.~Demetz, and J.~Weickert,
\newblock ``Why is the census transform good for robust optic flow
  computation,''
\newblock in {\em Scale Space and Variational Methods in Computer Vision},
  2013.

\bibitem{WuLimYang13}
Yi~Wu, Jongwoo Lim, and Ming-Hsuan Yang,
\newblock ``Online object tracking: A benchmark,''
\newblock in {\em IEEE Conference on Computer Vision and Pattern Recognition
  (CVPR)}, 2013.

\bibitem{mueller2016benchmark}
Matthias Mueller, Neil Smith, and Bernard Ghanem,
\newblock ``A benchmark and simulator for uav tracking,''
\newblock in {\em European conference on computer vision}. Springer, 2016, pp.
  445--461.

\bibitem{drone-tracking}
Siyi Li and Dit-Yan Yeung,
\newblock ``Visual object tracking for unmanned aerial vehicles: A benchmark
  and new motion models,''
\newblock in {\em AAAI}, 2017.

\bibitem{henriques2015high}
Jo{\~a}o~F Henriques, Rui Caseiro, Pedro Martins, and Jorge Batista,
\newblock ``High-speed tracking with kernelized correlation filters,''
\newblock {\em IEEE Transactions on Pattern Analysis and Machine Intelligence},
  vol. 37, no. 3, pp. 583--596, 2015.

\bibitem{danelljan2014accurate}
Martin Danelljan, Gustav H{\"a}ger, Fahad Khan, and Michael Felsberg,
\newblock ``Accurate scale estimation for robust visual tracking,''
\newblock in {\em British Machine Vision Conference, Nottingham, September 1-5,
  2014}. BMVA Press, 2014.

\end{thebibliography}

}
\end{document}